\documentclass{article}

\usepackage{arxiv}

\usepackage[utf8]{inputenc} 
\usepackage[T1]{fontenc}    
\usepackage{hyperref}       
\usepackage{url}            
\usepackage{booktabs}       
\usepackage{amsfonts}       
\usepackage{nicefrac}       
\usepackage{microtype}      
\usepackage{lipsum}		
\usepackage{graphicx}
\usepackage{natbib}
\usepackage{doi}
\usepackage{listings}  
\usepackage{float}  
\usepackage{xcolor} 

\usepackage{algorithm}
\usepackage{algpseudocode}
\usepackage{tabularx}
\usepackage{array}

\lstdefinelanguage{yaml}{
  morekeywords={true,false,null,y,n},
  sensitive=false,
  morecomment=[l]{\#},
  morestring=[b]"
}

\definecolor{codebg}{rgb}{0.97, 0.97, 0.97}  

\title{Small Dents, Big Impact: A Dataset and Deep Learning Approach for Vehicle Dent Detection}

\author{
    \href{https://orcid.org/0000-0000-0000-0000}{\includegraphics[scale=0.06]{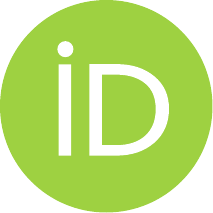}\hspace{1mm}Muhammad Danish Zia Baig} \\
    School of Electrical Engineering and Computer Science (SEECS) \\
    National University of Sciences and Technology (NUST) \\
    Islamabad, Pakistan \\
    \texttt{mbaig.msee21seecs@seecs.edu.pk} \\
    \And
    \href{https://orcid.org/0000-0002-3893-4330}{\includegraphics[scale=0.06]{orcid.pdf}\hspace{1mm}Mohsin Kamal} \\
    School of Electrical Engineering and Computer Science (SEECS) \\
    National University of Sciences and Technology (NUST) \\
    Islamabad, Pakistan \\
    \texttt{dr.mohsinkamal@seecs.edu.pk} \\
    \And
    \href{https://orcid.org/0000-0002-7330-6129}{\includegraphics[scale=0.06]{orcid.pdf}\hspace{1mm}Zahid Ullah} \\
    Dipartimento di Elettronica \\
    Informazione e Bioingegneria, Politecnico di Milano \\
    20133 Milan, Italy \\
    \texttt{zahid.ullah@polimi.it} \\
}

\renewcommand{\shorttitle}{\textit{arXiv} Template}

\hypersetup{
pdftitle={A template for the arxiv style},
pdfsubject={q-bio.NC, q-bio.QM},
pdfauthor={David S.~Hippocampus, Elias D.~Striatum},
pdfkeywords={First keyword, Second keyword, More},
}

\begin{document}
\maketitle

\begin{abstract}
    Conventional car damage inspection techniques are labor-intensive, manual, and frequently overlook tiny surface imperfections like microscopic dents.  Machine learning provides an innovative solution to the increasing demand for quicker and more precise inspection methods. The paper uses the YOLOv8 object recognition framework to provide a deep learning-based solution for automatically detecting microscopic surface flaws, notably tiny dents, on car exteriors.  Traditional automotive damage inspection procedures are manual, time-consuming, and frequently unreliable at detecting tiny flaws.  To solve this, a bespoke dataset containing annotated photos of car surfaces under various lighting circumstances, angles, and textures was created.  To improve robustness, the YOLOv8m model and its customized variants, YOLOv8m-t4 and YOLOv8m-t42, were trained employing real-time data augmentation approaches.  Experimental results show that the technique has excellent detection accuracy and low inference latency, making it suited for real-time applications such as automated insurance evaluations and automobile inspections. Evaluation parameters such as mean Average Precision (mAP), precision, recall, and F1-score verified the model's efficacy. With a precision of 0.86, recall of 0.84, and F1-score of 0.85, the YOLOv8m-t42 model outperformed the YOLOv8m-t4 model (precision: 0.81, recall: 0.79, F1-score: 0.80) in identifying microscopic surface defects.  With a little reduced mAP@0.5:0.95 of 0.20, the mAP@0.5 for YOLOv8m-t42 stabilized at 0.60.  Furthermore, YOLOv8m-t42's PR curve area was 0.88, suggesting more consistent performance than YOLOv8m-t4 (0.82).  YOLOv8m-t42 has greater accuracy and is more appropriate for practical dent detection applications, even though its convergence is slower. The visualizations of activation maps and anticipated bounding boxes further improve interpretability.  The suggested method provides a scalable and economical solution for minor dent identification, laying the groundwork for future research into 3D surface analysis and mobile platform deployment.
\end{abstract}


\section{Introduction}
Minor dents are one of the most commonly missed but serious types of surface damage found during a vehicle examination.  These dents, which are frequently shallow and irregular in appearance, pose a unique challenge for both human evaluators and automated systems because to their low contrast with adjacent surfaces and subtle geometric deformations.  Detecting faults in vehicles is crucial for insurance claims, quality control, and after-sales inspections, as even minor flaws can impact resale value and customer satisfaction \cite{thomas2023car} \cite{Raju2023}.  Traditionally, this activity has been conducted manually, relying on experienced specialists whose assessments might be subjective and inconsistent, particularly under changing lighting and surface conditions.

\begin{figure}[h]
	\centering
        \includegraphics[width=0.8\textwidth]{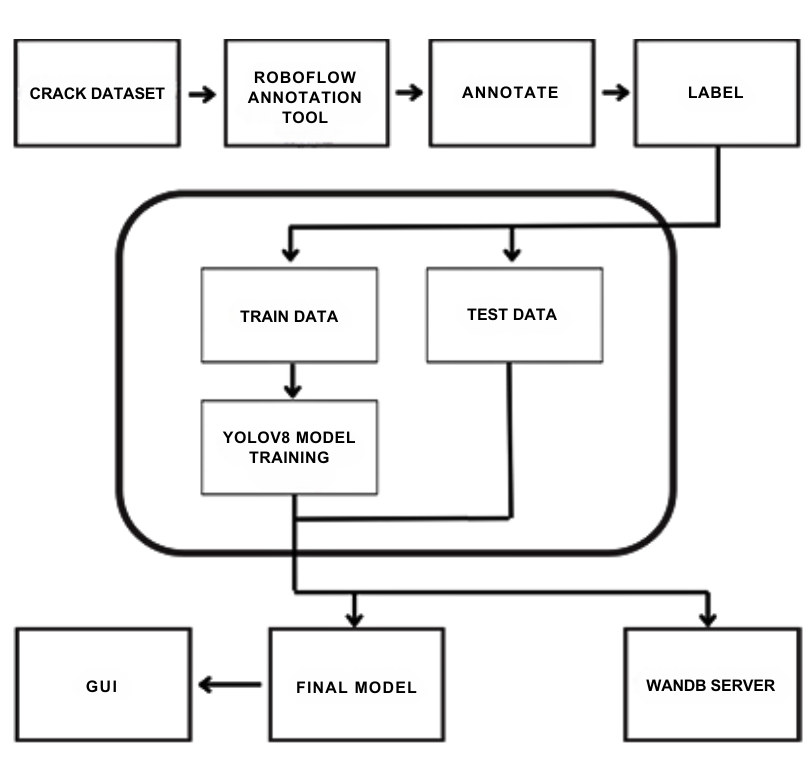}
	\caption{System Architecture of Car Damage Detection\cite{10915008}.}
	\label{fig:fig1}
\end{figure}

To address this demand, deep learning-based object identification models, particularly those from the YOLO (You Only Look Once) family, have proven to be exceptionally effective. YOLO models are particularly valued for their ability to strike a balance between detection accuracy and speed, making them ideal for real-time applications in industrial and commercial environments. The most recent version, YOLOv8, offers significant improvements in detection speed, model correctness, and flexibility in a wide range of complex circumstances. A recent study \cite{10915008} employed YOLOv8 to build a real-time vehicle damage detection system, as demonstrated in Figure.~\ref{fig:fig1}.   The dataset used to train this system classified eight different types of automobile damage, including shattered windshields, dents, and damaged bumpers.   This foundation provides an excellent starting point for more specific research into identifying smaller types of damage, such as small dents, which requires even more accurate feature extraction and classification.

\begin{figure}
	\centering
        \includegraphics[width=0.8\textwidth]{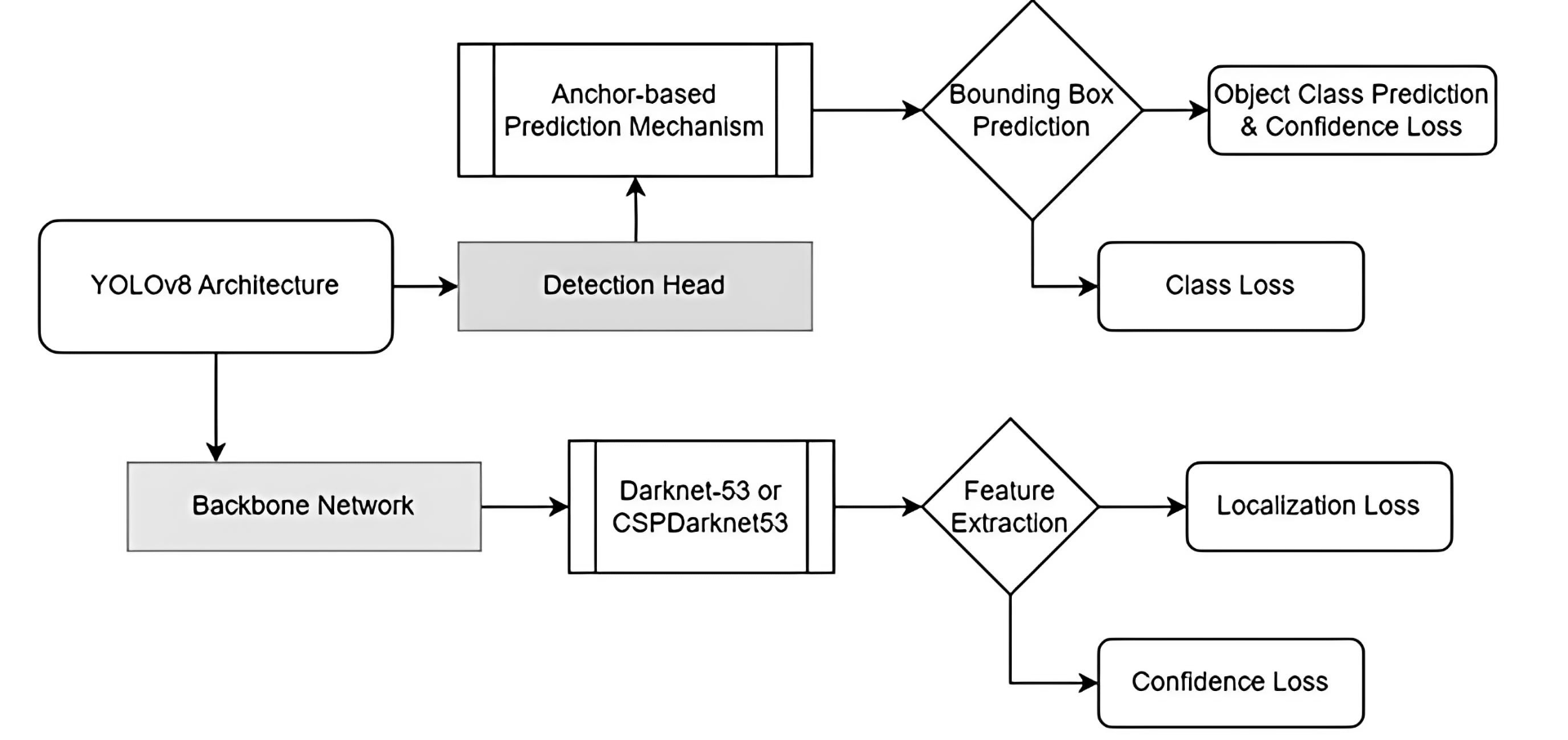}
	\caption{System Architecture of Car Damage Detection\cite{10915008}.}
	\label{fig:fig2}
\end{figure}

Recent improvements in deep learning have resulted in highly specialized models that can detect even minor and sparse surface flaws. \cite{zhang2023yolov5} improved the YOLOv5 architecture by incorporating a BiFPN-based feature fusion technique and an extra small-object detection layer, resulting in significant gains in detecting minute faults on metallic surfaces.  Although their study is not vehicle-specific, the technical issues they addressed are similar to those found in automobile dent detection.  Similarly, \cite{Agrawal2025} developed a YOLOv8-based hybrid model that can categorize and segment numerous damage types shown in Figure.~\ref{fig:fig2}, including dents, across several vehicle sections, highlighting the versatility of such architectures.

Transfer learning has also gained popularity in this field. \cite{sruthy2021transfer} fine-tuned CNN models such as ResNet50 and VGG16 using labeled vehicle damage datasets.  Their approach enabled pre-trained models to adapt to specific visual patterns of car damage, such as tiny dents.  They did, however, notice constraints caused by dataset imbalance and insufficient segmentation granularity, both of which are common concerns with dent detection methods.  These findings highlight the need for more refined models and diverse training datasets that can generalize across a variety of environmental conditions.

To meet real-time processing constraints, researchers have optimized lighter variants of YOLO, such as YOLOv7-tiny and YOLOv7x for automotive damage detection \cite{ramazhan2023yolov7}.  When trained on annotated photos of scratches, dents, and shattered glass, these variations achieved strong F1-scores, with YOLOv7x reaching 0.949.  The use of image augmentation techniques, which improved robustness to lighting and orientation variability, increased the effectiveness of such models even more.  However, their ability to detect microscopic, low-contrast dents remains strongly dependent on dataset quality and augmentation approach.

A growing amount of research has begun to use segmentation and detection approaches for greater performance in difficult circumstances. \cite{Bhatlawande2023} introduced a hybrid pipeline that uses YOLOv5 for object localization and watershed segmentation for contour extraction.  This method allows for fine-grained recognition of dent shapes, making it ideal for shallow and mild surface deformations.  Other studies, such as those by \cite{Wang2023} and \cite{Chua2021}, have introduced methods such as structured light imaging and modular CNN-based classifiers that identify distortions on reflective surfaces and distinguish specific auto parts and their damage status.

Several recent studies have attempted to address the difficulty of dent detection using a variety of methodologies. \cite{polyantseva2025} developed a method for localizing deformations using Laplacian-based preprocessing and unsupervised K-Means clustering, which proved effective in controlled situations.  However, their method failed with generalization under different illumination and background textures. \cite{widjojo2022} used deep CNN-based approaches, specifically localized patch learning and transfer learning, to greatly enhance the classification of visible dents.  While these models performed admirably in detecting obvious surface defects, their sensitivity to tiny, shallow, or shaded dents remained a shortcoming.  Collectively, these findings highlight the need for more universal and resilient architectures—such as YOLOv8 and transformer-based models—that can capture minute visual cues across a wide range of real-world scenarios.

Despite these developments, a substantial barrier persists: a lack of high-quality, annotated datasets focusing on tiny dents.  Generic datasets for vehicle damage identification, such as CarDD \cite{Wang2023CarDD} and VehiDE \cite{Huynh2023VehiDE}, frequently have few examples of subtle faults.  This absence of representation reduces model generalization and training effectiveness.  Researchers such as \cite{widjojo2022} and \cite{polyantseva2025} advocate for domain-specific dataset preparation, including multi-view and augmented samples, to increase performance on less visible damage types.

In light of these findings, this paper seeks to offer a deep learning-based model optimized for minor dent identification, with an emphasis on improving robustness to changeable lighting, surface reflection, and dent shape.  The suggested approach builds on recent architectural advances in models like YOLOv8 \cite{10915008} and LCG-YOLO \cite{yu2024}, leveraging real-time object detection techniques, data augmentation, and explainability tools like Grad-CAM.  The ultimate goal is to create an efficient, precise, and scalable dent detection system that can help with insurance automation, smart maintenance, and industrial quality assurance jobs requiring dependability and speed. The following are this work's key contributions:

\begin{itemize}
    \item Using the YOLOv8m architecture, a deep learning-based pipeline was created to identify and locate small dents on automobile bodies, concentrating on minute surface irregularities that are frequently overlooked by traditional techniques.
    \item To better replicate real-world settings, a difficult dataset of annotated automobile photos with small dents taken from various perspectives and lighting conditions was created and curated.
    \item To overcome data scarcity and enhance generalization across various vehicle surfaces and backdrops, data augmentation techniques and adjusted model hyperparameters were implemented.
    \item The model's efficacy and dependability in minor dent identification tasks were proved by a thorough examination utilizing standard metrics (precision, recall, F1-score, and mAP).
    \item In order to enhance the interpretability of the model and facilitate its incorporation into actual automotive inspection systems, visualization techniques including confidence maps and bounding box overlays were used.
\end{itemize}

The study problem, significance, goals, and constraints pertaining to minor dent identification are described in Section 1.  The relevant research on automated vehicle damage identification is reviewed in Section 2.  Preprocessing and dataset collection are covered in Section 3.  The detection mechanism based on YOLOv8 is explained in Section 4.  Experimental results are shown in Section 5 along with a visual analysis and quantitative metrics.  The paper's conclusion and recommendations for future research are made in Section 6.

\section{Literature Review}
\label{sec:literature}
YOLO designs' speed and localization precision have made them useful for detecting damage in automobiles. \cite{10493214} detected fractures and dents on concrete and metallic surfaces using YOLOv3 with MATLAB preprocessing.  Their strategy combined the Chan-Vese method for picture segmentation with an emphasis on low-cost Arduino deployment.  However, possible overfitting was indicated by a notable discrepancy in accuracy between training (70\%) and validation (45\%).  With a mAP of 82.2\%, \cite{Meenakshi2023} developed a web system based on YOLOv3 for categorizing 21 damage kinds.  Due to dataset restrictions, performance decreased on minor or less obvious faults, yet being extremely accurate on discrete categories like hood dents.

\cite{xu2021modifiedyolo} improved the identification of small-scale metal defects by enhancing YOLOv3 with extra feature layers and K-Means++ clustering.  Despite persistent misclassifications on textured surfaces, their model maintained real-time speed (83 FPS) and obtained 75.1\% mAP.  With a tiny object layer and BiFPN fusion, \cite{zhang2023yolov5} suggested YOLOv5-HA, increasing mAP to 95.3\% on NEU-DET.  The study lacked assessments under different lighting and occlusion conditions, but showing promise for minor flaws.

To identify damage in the actual world, \cite{ramazhan2023yolov7} compared YOLOv7 variations.  With an F1-score of 0.949, YOLOv7x demonstrated inadequate efficiency on edge devices and lacked severity rating. \cite{Agrawal2025} achieved 88.2\% mAP by combining YOLOv8-based classification and segmentation.  Subtle defects caused performance to decline, indicating continued difficulties with low-contrast dent detection.  \cite{Raju2023} used YOLOv8 with a CSPDarknet53 backbone and a self-attention detection head.  Their Flask and ReactJS-deployed system produced a maximum mAP of 0.543 and 155 frames per second.  However, under occlusion, performance decreased and scratch detection delayed.

Using YOLOv10 with a five-camera configuration and optimal lighting, \cite{polyantseva2025} achieved a mAP50 of 0.870.  It performed poorly on small dents and mobile inspections, despite being efficient in controlled settings. \cite{Luo2023} enhanced YOLOX for the detection of unusual defects by adding a bespoke focus loss and decoupled head.  The model demonstrated strength in detecting wrinkles and pinholes, achieving 91.5\% mAP, suggesting promise for tiny dent identification.  For improved texture and scale sensitivity, \cite{yue2023metal} presented Metal-YOLOX with TGSR and HCNet modules.  It was accurate (up to 81.2\% mAP) and lightweight, making it suitable for edge deployment; nonetheless, it lacked resilience under changeable lighting and dent-specific evaluation.

To differentiate between surface scratches and structural dents, a Mask R-CNN model was employed in conjunction with structured light and dual-focus cameras.  The method worked consistently with different materials and lighting conditions, but it was not scalable for large-scale or real-time industrial applications~\cite{Wang2023}.  A three-stage framework was presented, which uses Mask R-CNN to segment damage and EfficientNet and MobileNetV2 to classify severity.  The F1-score improved by 9\% once damage masks were included.  However, the model's ability to detect subtle problems such tiny dents was diminished by the restricted and unbalanced dataset~\cite{widjojo2022}.

ResNet50 achieved a greater mAP (0.14 vs. 0.11) in the Mask R-CNN architecture than ResNet101, according to a comparative research on backbone networks.  The shallower network is better suited for identifying small-scale flaws since it is more successful at concentrating on specific damage areas~\cite{shakhovska2023}.  For the detection of fender defects, an R-CNN model improved with an Adaptive Genetic Algorithm and supplemented by Mach band illumination was suggested.  Although its assessment on synthetic data limited its generalizability, it achieved 98.5\% accuracy and a 13.7-pixel localization error~\cite{park2020rcnn}.

VGG16 and ResNet50 were used in a multi-model transfer learning strategy that produced an 87.9\% classification accuracy for damage intensity.  Although the model performed well, it was less reliable for detecting tiny dents because it was dependent on a small, manually labeled dataset~\cite{sruthy2021transfer}.  Another hybrid approach achieved a 77.8\% mAP by combining YOLO for object localization and pre-trained CNNs for feature extraction.  Its poor feature sensitivity made it difficult to detect fine-grained defects, even though it was appropriate for apparent damage~\cite{Dwivedi2021}.  Through affine augmentation, a method that used CNNs for feature extraction and SVM for classification reported 94\% precision and adaptability to changes.  Its accuracy in identifying small dents was still restricted, despite its effectiveness in identifying overall damage~\cite{verma2023car}.

Recent developments in automatic dent identification use a combination of traditional image processing and deep learning.  Using a dataset of 831 enhanced pictures, \cite{Bhatlawande2023} improved tiny dent border localization by combining YOLOv5 with watershed segmentation.  Although it worked well in a variety of illumination conditions, gradient-based segmentation resulted in some areas being overlooked. \cite{zhou2020} suggested a hybrid method for inspecting metal surface defects that combines BEMD decomposition and Canny edge identification.  BEMD increased signal-to-noise ratios, but it was responsive to surface orientation and required a lot of processing. In 3D methods, \cite{Lilienblum2000} detected deformations as small as $20~\mu m$ using a photogrammetric system with neural networks.  The technology needed preset settings and lacked real-time performance despite its accuracy.  Using 767 annotated photos, \cite{setyawan2023yolo} built YOLOv5 for multi-class damage detection.  Under varying lighting conditions, their best model (YOLOv5x) helped detect dents with mosaic augmentation, achieving an F1-score of 0.908.

Model training and assessment are greatly aided by publicly available datasets like VehiDE \cite{Huynh2023VehiDE} and CarDD \cite{Wang2023CarDD}.  Whereas VehiDE delivers 13,945 high-resolution photos with 32,000+ cases, CarDD offers 9,000+ damage annotations, 38.6\% of which are small-scale.  The necessity for specific methods is shown by the fact that models still have trouble addressing minute flaws like dents, even with the abundance of datasets. As demonstrated in \cite{shakhovska2023}, custom datasets that have been curated with expert input address class balancing across damage severity levels and domain-specific labeling.  In a similar vein, \cite{Patil2017} stressed the use of customized datasets and transfer learning for fine-grained classification of microscopic flaws such as tiny scratches. \cite{Tang2023} emphasized the importance of classical pre-processing (such as wavelets, LBP, and Gabor filters) in exposing subtle flaws that apply to small dents.  The review offers a structured pipeline pertinent to vehicle surfaces, notwithstanding its steel concentration.

Low contrast and resolution continue to be major problems.  Due to limited samples and poor image quality, \cite{Chua2021} found considerable overfitting and poor validation accuracy (49.28\%), particularly for damage classification.  Furthermore, \cite{Fouad2023} shown that inconsistent annotations and fixed low-resolution photos (150$\times$150 px) hinder the detection of minor dent.  The benefits of YOLOv5x were illustrated by \cite{setyawan2023yolo}; nevertheless, their absence of dent-specific augmentation and grouping of tiny dents with broader classes hampered precision for shallow deformations. For nuanced patterns, more sophisticated techniques like as GANs, ResNets, and transformers provide superior generalization \cite{Tang2023}. \cite{Patil2017} noted that minor dents are frequently mistaken for "no damage" classes because of their poor visual cues, and they reached an accuracy of 89.5\% using transfer learning and composition techniques.  Although controlling occlusion and dataset variety continued to be difficulties, their sliding window method with softmax posteriors assisted in localizing fine damage.

Using attention mechanisms and improved feature fusion, recent research enhanced YOLO-based models for defect identification.  With the addition of CBAM and SIoU, \cite{yu2024} improved YOLOv5, attaining high precision but without dent-specific tweaking.  By employing high-resolution layers and enhanced anchor clustering, \cite{xu2021modifiedyolo} increased YOLOv3's sensitivity to minor defects.  In order to improve subtle defect detection, \cite{9870791} enhanced YOLOv4 using fusion blocks and intelligent augmentation.  YOLOv7-Tiny was implemented on edge devices by embedded solutions such as \cite{qin2023yolov7}, providing real-time performance.  Despite their strong overall results, YOLOv8-based investigations \cite{Agrawal2025, Raju2023, 10915008, ijcrt2024car} still have trouble identifying minor, low-contrast dents.  Bounding boxes are used by most, which restricts the localization of fine damage.  Using a YOLOv8m-based system designed for microscopic dent detection, this paper seeks to close that gap.

\begin{table*}[h]
\caption{Critical Analysis of YOLOv8-Based Dent Detection Studies}
\centering
\scriptsize
\renewcommand{\arraystretch}{1.4} 
\setlength{\tabcolsep}{4pt}       
\begin{tabular}{c|c|c|c|c}
\hline
\textbf{Attribute} & \textbf{\cite{Agrawal2025}} & \textbf{\cite{Raju2023}} & \textbf{\cite{10915008}} & \textbf{\cite{ijcrt2024car}} \\
\hline
Dataset & 4k+ HR images & 465 Kaggle imgs & 485 labeled imgs & Not specified \\
Method & YOLOv8 + seg/class & YOLOv8 + CSP + attn & YOLOv8 + SGD + RFlow & YOLOv8 + TL + mask \\
Strength & mAP 88.2\%, 8-class & 155 FPS, GUI & F1=0.92, GUI & Pixel-level, real-time \\
Limitation & Weak on fine/low-contrast & Low scratch mAP, lighting & Small set, lacks fine detail & No eval on faint/overlap dents \\
Relevance & Shows limits on fine damage & RT GUI, small features & Baseline F1, no faint test & Closely aligned \\
\hline
\end{tabular}
\label{tab:yolov8_summary}
\end{table*}

YOLOv8 provides precise, quick, and deployable solutions for detecting vehicle damage as shown in Table \ref{tab:yolov8_summary}.  Most models have trouble detecting small, subtle, or overlapping dents in complex lighting or occlusion, but they are good at detecting larger flaws like dents and scratches.  Transfer learning, dataset quality, segmentation accuracy, and implementation strategies are important considerations.  The purpose of this work is to address YOLOv8's shortcomings in terms of micro-level dent detection.

\section{Dataset Curation}
\label{sec:dataset}
The foundation of any successful computer vision project is the dataset's quality, applicability, and organization.   The problem of detecting small dents on car surfaces has few publicly available datasets, especially those that concentrate on limited, subtle dents from different angles and in different lighting circumstances.   To overcome this limitation and give the research community a valuable resource, a new, self-collected dataset was created and arranged using robust annotation techniques. The dataset used for training and evaluation is published by M. D. Z. Baig as "Minor Dents Detection Dataset for Car Body Panels" and is publicly available on Zenodo\cite{dataset_danish}.

\subsection{Dataset Motivation}
Tiny dents are among the most common yet often disregarded types of auto damage.   Tiny dents may be harder to spot than larger collisions or deformations due to their shallow depth, irregular shape, and appearance in various light conditions.   Deep learning-based automated dent detection requires a dataset that captures this flexibility.   Since there aren't many datasets with minor dent annotations available, a custom dataset was created to address this problem.   The goal was to produce a high-quality dataset with diverse, realistic samples that reflected actual circumstances that occur in quality control, insurance, and vehicle inspection contexts.

\subsection{Data Collection}
\begin{figure}
	\centering
        \includegraphics[height=12cm]{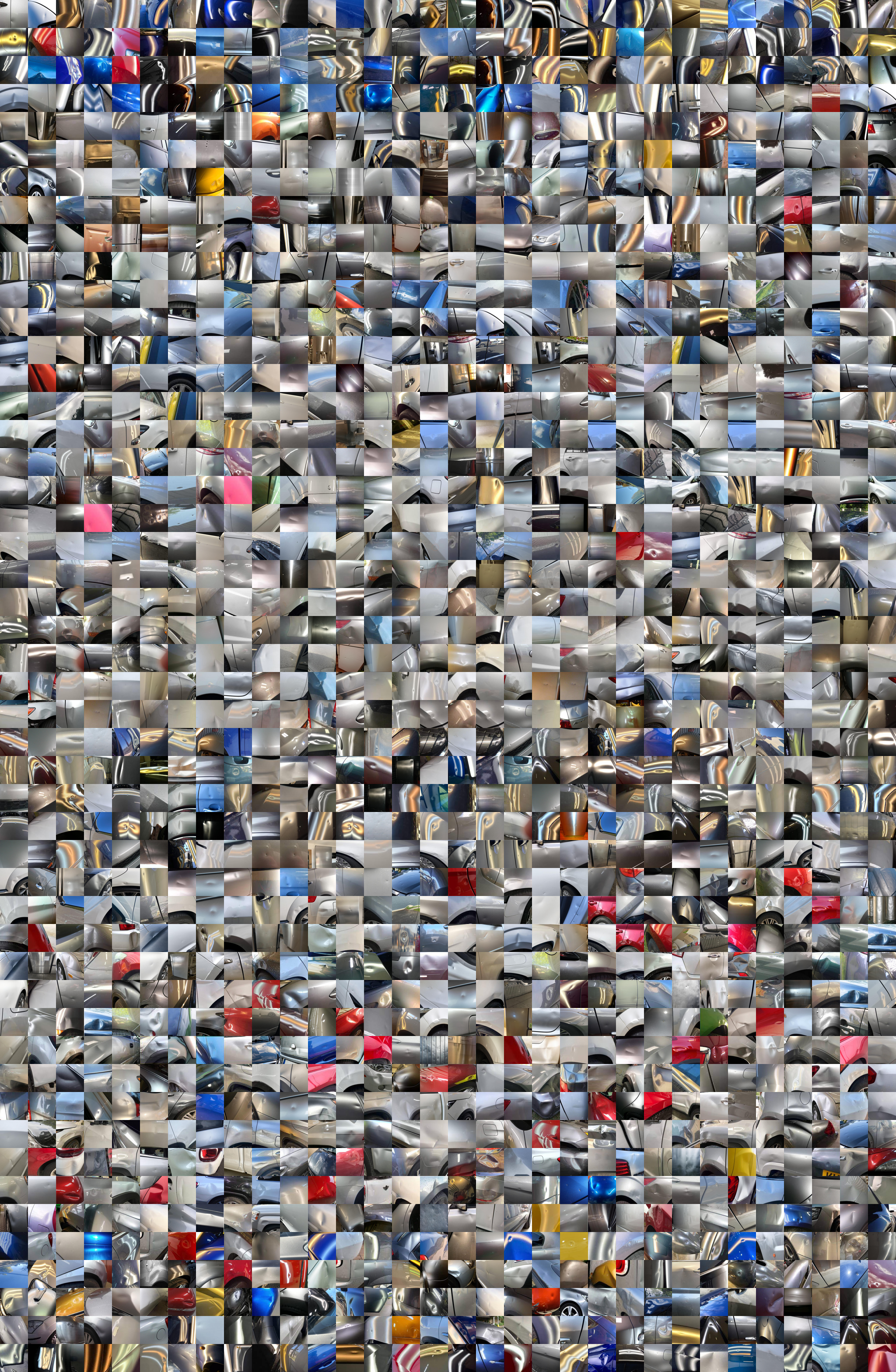}
	\caption{Collected Dataset Training Images.}
	\label{fig:fig3}
\end{figure}

To collect the photos for this dataset, a simple iPhone camera was utilized, which offers a realistic and usable image quality level for practical applications.   In addition, the dataset's usage of a consumer-grade mobile device rather than specialized equipment supports its relevance to real-world deployment situations, where car inspection systems may operate in uncontrolled circumstances.   Both indoor and outdoor settings, including workshops, public parking lots, private garages, and road conditions, were utilized to gather the data. These settings included natural fluctuations in lighting, shadows, and surface reflections—all of which are critical for real-world dent detection.  The 2,241 photos in the collection guaranteed a thorough representation of car surfaces (doors, bumpers, fenders, and hoods) with tiny dents.   Each image captured subtle imperfections that are often difficult to see due to their surface-level nature and uneven look. The training images are shown in Figure~\ref{fig:fig3}.

\subsection{Dataset Structure}
The dataset is structured into three distinct subsets: training, validation, and test. Each subset contains an \texttt{images/} directory and a \texttt{labels/} directory, conforming to the format used by the YOLOv8 object detection framework.

The training set comprises a total of 1,568 images in \texttt{.jpg} format and 1,569 label files in \texttt{.txt} format. Each label file corresponds to an individual image and contains bounding box annotations in YOLO format. However, one additional label file was identified, which suggests a possible mismatch or a redundant label. This discrepancy is currently under review to ensure annotation consistency.

The validation set consists of 449 images and 449 corresponding label files, with a one-to-one match between images and labels. This subset is used during training to monitor the model’s performance and prevent overfitting.

The test set includes 224 images and 224 label files, again with a one-to-one correspondence. This portion of the dataset is reserved for evaluating the final trained model to assess its generalization capability on unseen data.

Each label file corresponds to one image file with the same base name and contains object annotations in YOLOv8 format. The annotations include the class index followed by the normalized bounding box coordinates: (class\_id, x\_center, y\_center, width, height), where all values are normalized between 0 and 1 relative to the image dimensions.

A summary of the dataset split and file distribution is provided in Table~\ref{tab:dataset_summary}, which outlines the number of images and label files in each subset.

\begin{table}[h]
\centering
\caption{Summary of Dataset Subsets}
\begin{tabular}{c|c|c}
\hline
\textbf{Subset} & \textbf{Images (.jpg)} & \textbf{Labels (.txt)} \\
\hline
Training        & 1,568                   & 1,568                  \\
Validation      & 449                     & 449                     \\
Test            & 224                     & 224                      \\
\hline
\end{tabular}
\label{tab:dataset_summary}
\end{table}

\subsection{Annotation Process}
The Roboflow platform, which offers a user-friendly interface for drawing bounding boxes, was used to complete the annotating process.  To ensure that only modest, localized deformations were labeled, annotators were told to solely concentrate on minor dents that were readily visible.  Unless they co-occurred with a tiny ding of relevance, larger dents, scratches, or paint discolorations were not included.  Tight bounding boxes were prioritized in order to improve detection performance and lower background noise.  To represent the binary classification goal of separating dents from the background, a single class label was applied across the dataset.

\begin{figure}[h]
    \centering
    \includegraphics[height=12cm]{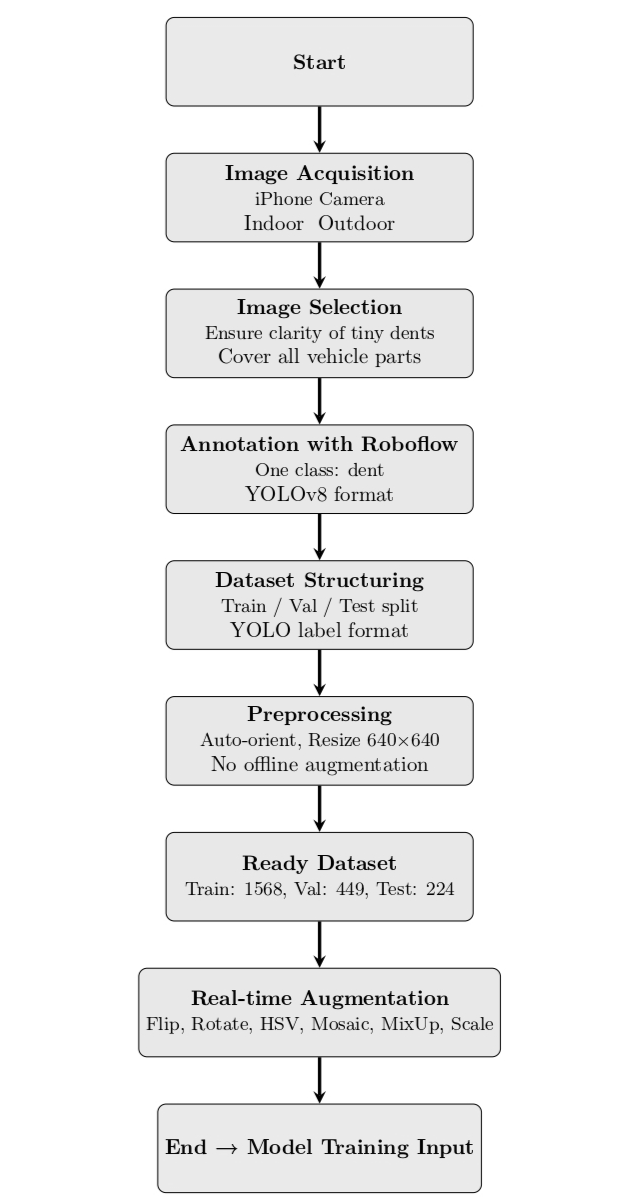}
    \caption{Dataset curation block diagram.}
    \label{fig:fig4}
\end{figure}

\subsection{Dataset Preprocessing}
The Roboflow platform was used to apply a number of preprocessing processes in order to get the dataset ready for training using cutting-edge object detection models:
\subsubsection{Auto Orientation}
To guarantee consistency, embedded EXIF metadata was used to automatically align every image.  This assisted in getting rid of irregularities brought on by the iPhone camera's rotation settings.
\subsubsection{Resizing}
To make sure they would operate with contemporary object recognition frameworks, especially YOLOv8, all photos were scaled to the usual 640$\times$640 pixel resolution.  Since stretch interpolation was used for this resizing, the original aspect ratio was not maintained.  By standardizing the input dimensions in this way, YOLO's neural architecture can operate at its best, allow uniform input formatting for batch training, and save GPU memory and training time.
\subsubsection{No Augmentation}
During the whole dataset production process, no artificial augmentations (such as flipping, cropping, rotation, brightness fluctuation, or Gaussian noise) were employed.   The decision was made to preserve the authenticity and realism of each image so that the model can learn from naturally occurring differences.   Data augmentation was subsequently added dynamically during the training process.
\subsubsection{Real-Time Augmentation (During YOLO Training)}
Real-time data augmentation was widely used during the training phase utilizing the YOLOv8 framework, even though no augmentations were added to the raw dataset during preprocessing.  This approach dynamically increases input diversity during runtime while maintaining the integrity of the original dataset on disk.  Similar to earlier iterations of YOLO, YOLOv8 facilitates effective on-the-fly augmentation, which reduces overfitting and enhances the model's ability to generalize to new data. Transformations such random horizontal flips, scaling, zooming, rotations, HSV changes (hue, saturation, brightness), translations, cropping, mosaic augmentation (combining four pictures), and MixUp (combining two images and labels) are added to each batch during training.  The robustness of the model is increased by these transformations, which mimic real-world variations such as shifts in viewpoint, lighting, partial occlusions, and camera movement.  In addition to keeping the dataset clean and consistent, this real-time method exposes the model to a variety of synthetic scenarios throughout each epoch, greatly enhancing its capacity to identify small dents on a variety of automotive surfaces and environmental circumstances.

\subsection{Annotation Format}
Using Roboflow, bounding boxes were manually drawn around only small visible dents (single class). Labels follow the YOLO format:

\texttt{<class\_id> <x\_center> <y\_center> <width> <height>}

All values are normalized relative to 640$\times$640 resolution.

\subsubsection{YAML Configuration}
\begin{figure}[h]
\centering
\begin{minipage}{\linewidth}
\lstset{
  language=yaml,
  basicstyle=\ttfamily\footnotesize,
  keywordstyle=\color{blue},
  commentstyle=\color{gray},
  stringstyle=\color{orange},
  showstringspaces=false,
  breaklines=true,
  columns=flexible,
  backgroundcolor=\color{codebg},
  frame=single,
  rulecolor=\color{gray},
  captionpos=b,
  aboveskip=0pt,
  belowskip=0pt
}
\begin{lstlisting}[caption={YOLOv8 Dataset Configuration}, label={lst:data_yaml}]
train: ./train/images
val: ./valid/images
test: ./test/images

nc: 1
names: ['dent']
\end{lstlisting}
\end{minipage}
\vspace{-1em}
\end{figure}

\subsection{Novelty and Contributions}
This dataset presents a genuine, useful problem by highlighting the detection of subtle, small dents under various lighting conditions.   The stages of the dataset curation are shown in Figure \ref{fig:fig4}. This sets it apart from artificial or studio-quality datasets in that it uses a nonprofessional iPhone camera to add a substantial sense of realism and accessibility.   This also makes it more appropriate for edge devices and mobile deployment scenarios.  Additionally, due of its YOLO-standard form (640x640 resolution and normalized annotations), researchers and practitioners can easily use this dataset for training, benchmarking, or transfer learning in object recognition tasks.

\section{Methodology}
\label{sec:methodology}
The YOLO (You Only Look Once) object detection family has evolved dramatically over time.  The groundwork for real-time object recognition was laid by YOLOv1 through YOLOv3, with YOLOv3 including residual connections and multi-scale predictions \cite{ultralytics_models}.  By utilizing cutting-edge training techniques, Cross-Stage Partial (CSP) connections, and mosaic augmentation, YOLOv4 further improved accuracy.  Ultralytics created YOLOv5 in PyTorch, and because of its versatility, functionality, and thorough documentation, it gained widespread use.  The goal of YOLOv6 and YOLOv7 was to maximize speed and efficiency for edge and industrial deployments.  Lastly, Ultralytics completely redesigned YOLOv8, providing cutting-edge performance and support for a variety of vision tasks \cite{ultralytics_yolov8}.

\subsection{YOLOv8 Model: Selection and Variants}

Ultralytics' latest vision model, YOLOv8, enables classification, instance segmentation, object identification, and posture estimation.   Unlike previous versions, YOLOv8 has been entirely revamped with a modern architecture, eliminating any remaining drawbacks from previous YOLO implementations.

\subsubsection{Selection of YOLOvX}
YOLOv8 was selected due to its combination of state-of-the-art accuracy, inference speed, and deployment flexibility. Unlike earlier versions, YOLOv8 is a complete rewrite with modern components, allowing it to outperform legacy models in both standard benchmarks and real-world scenarios. Its anchor-free head, modular architecture, and native support for multi-task learning (detection, segmentation, classification, and pose estimation) make it particularly suitable for detecting subtle and small objects like minor dents in vehicle surfaces.

With a number of improvements, YOLOv8 is a strong option for object detection jobs.  Its anchor-free design makes training easier and enhances the ability to recognize things with irregular shapes, such as dents.  Its dynamic input processing improves robustness in a range of situations by accommodating different image sizes and aspect ratios.  The model is appropriate for edge deployment since it can export to formats like ONNX, TensorRT, and CoreML across platforms.  Furthermore, YOLOv8 improves generalization by providing native support for sophisticated augmentation methods like MixUp and Mosaic.  Supported by a vibrant ecosystem that Ultralytics keeps up to date, it gains from frequent upgrades, robust community support, and thorough documentation.

\subsubsection{YOLOv8 Model Variants}
YOLOv8 is available in five model scales. These differ in the number of layers and parameters as depicted in Table \ref{tab:yolov8_models}
\begin{table}[h]
\centering
\caption{Comparison of YOLOv8 Model Variants by Parameters, Speed, and Accuracy\cite{ultralytics_yolov8}}
\label{tab:yolov8_models}
\begin{tabular}{c|c|c|c}
\hline
\textbf{Model} & \textbf{Parameters} & \textbf{Speed} & \textbf{Accuracy (mAP@50)} \\
\hline
YOLOv8n & \textasciitilde3M & Fastest & Lower \\
YOLOv8s & \textasciitilde11M & Very Fast & Good \\
\textbf{YOLOv8m} & \textasciitilde25M & Balanced & \textbf{High} \\
YOLOv8l & \textasciitilde50M & Slower & Higher \\
YOLOv8x & \textasciitilde68M & Slowest & Highest \\
\hline
\end{tabular}
\end{table}
For this research, the YOLOv8m variant was selected as the base model due to its balance between accuracy and speed.

\subsubsection{Selected Model Variant: YOLOv8m}
The YOLOv8m model was selected from among the five YOLOv8 variations (n, s, m, l, x) due to its ideal trade-off between detection accuracy, inference time, and model complexity.  It has about 25 million parameters, making it both powerful enough to guarantee high precision and lightweight enough for real-time applications.  YOLOv8m is ideally suited for deployment on edge devices and consistently delivers performance on mid-range GPUs.  Its choice was further supported by preliminary testing on dent detection datasets, where the model outperformed smaller variants in detecting minute damage features.

\subsection{Procedure}
\begin{figure}[h]
    \centering
    \includegraphics[width=1\textwidth]{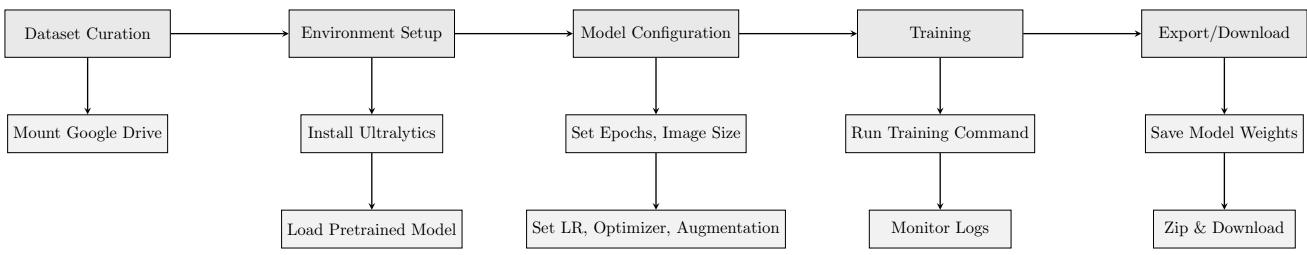}
    \caption{YOLOv8 Dent Detection Pipeline.}
    \label{fig:fig5}
\end{figure}

The underlying \textbf{YOLOv8m} model was used in two separate training sessions to assess the effectiveness of the suggested strategy under various training circumstances.  The Ultralytics YOLOv8 framework was used to carry out these training runs, and it automatically created the experiment directories with the names \texttt{YOLOv8-t4} and \texttt{YOLOv8m-t42}.  It should be noted that these names refer to different experiment folders that contain the training outputs, including weight files, setup settings, logs, and evaluation charts, rather than different architectures or model variations. As shown in Figure \ref{fig:fig5}, the YOLOv8 Dent Detection Pipeline outlines the steps for effectively detecting dents on car surfaces.

Using common augmentation methods like Mosaic, MixUp, and HSV shift, the model was trained for 100 epochs on the minor dent detection dataset in the \texttt{YOLOv8-t4} experiment.  The best-performing weights were saved with other files like \texttt{args.yaml} and visual performance charts, and the model was tracked during training based on validation loss and mean average accuracy at 0.5 IoU (mAP@0.5). Although the \texttt{YOLOv8m-t42} experiment was likewise run over 100 epochs, it might have included minor adjustments to the augmentation probabilities or hyperparameters.  A distinct set of evaluation metrics and plots were generated by this run, and these were then utilized to compare performance, especially with regard to generalization ability and edge case detection accuracy.  Notwithstanding these distinctions, the same YOLOv8m architecture was used in all trials; the only things that differed were session-specific setups, random seeds, or tuning parameters.

\subsection{Algorithm: YOLOv8 Training Pipeline}

YOLOv8 model training for dent identification on automobile surfaces is described in Algorithm 1.  Pretrained weights, a configuration file (\texttt{data.yaml}), and predetermined training parameters are fed into the pipeline at the start.  To access the dataset, external storage (like Google Drive) must first be mounted, and necessary libraries, such \texttt{ultralytics}, must be installed.  The supplied weights, $W_0$, are then used to load the pretrained YOLOv8 model.

Important training parameters are then established, such as a 640x640 pixel image size, a batch size of 16, and 100 training epochs.  The optimizer, Stochastic Gradient Descent (SGD), is employed. The learning rate is initially set at $lr_0 = 0.01$ and ends up being $lrf = lr_0 \times 0.1$.  Throughout the training, the learning rate is modified using a cosine learning rate schedule.  Furthermore, methods for data augmentation are made possible to improve model generalization.

Using the configurations listed in the \texttt{data.yaml} file, training continues.  The model weights and logs are saved to an output directory for subsequent analysis or deployment once training is finished.  The final results can be downloaded and are compressed, allowing for real-time application or additional examination.

\begin{algorithm}[h]
\caption{YOLOv8-m Dent Detection Training}
\begin{algorithmic}[1]
\State \textbf{Input:} Pretrained weights $W_0$, configuration file \texttt{data.yaml}, training parameters
\State \textbf{Output:} Fine-tuned YOLOv8 model for dent detection

\Procedure{TrainDentDetector}{$W_0$, \texttt{data.yaml}, params}
    \State Mount external storage (e.g., Google Drive) to access dataset
    \State Install required libraries: \texttt{ultralytics}
    \State Load pretrained YOLOv8 model using weights $W_0$
    \State Define training hyperparameters:
    \State \hspace{1em}Image size: 640$\times$640
    \State \hspace{1em}Batch size: 16
    \State \hspace{1em}Epochs: 100
    \State \hspace{1em}Optimizer: SGD
    \State \hspace{1em}Initial learning rate: $lr_0 = 0.01$
    \State \hspace{1em}Final learning rate: $lr_f = lr_0 \times 0.1$
    \State \hspace{1em}Cosine learning rate schedule
    \State \hspace{1em}Data augmentation enabled
    \State Start training using configuration in \texttt{data.yaml}
    \State Save model weights and training logs to output directory
    \State Zip and download the results
\EndProcedure
\end{algorithmic}
\end{algorithm}

\subsection{Performance Metrics}

To evaluate the performance of the YOLOv8m object detection model for minor dent detection, a comprehensive set of metrics was used to capture both classification accuracy and localization precision. Key evaluation indicators include Precision, Recall, F1-Score, Intersection over Union (IoU), mean Average Precision (mAP), and the Confusion Matrix.

The Confusion Matrix provides a summary of the model's prediction outcomes, distinguishing between true positives (correctly detected dents), false positives (incorrect dent predictions), false negatives (missed dents), and true negatives (correctly identified non-dent areas, though rare in object detection tasks). From this matrix, Precision and Recall are derived: precision evaluates the proportion of correct positive predictions, while recall assesses the ability to detect all relevant dents. These metrics are further combined into the F1-Score, which balances the trade-off between precision and recall, especially useful in imbalanced datasets like ours.

For object localization, Intersection over Union (IoU) measures the overlap between predicted and actual bounding boxes, with a threshold (commonly IoU $a \ge b$ 0.5) determining true positives. The mean Average Precision (mAP), particularly mAP@0.5, aggregates performance across different recall thresholds to provide a holistic view of detection capability.

Additionally, YOLOv8’s loss function integrates multiple components: Box Loss (IoU-based) penalizes inaccurate bounding box predictions, Class Loss captures misclassification errors, and Distribution Focal Loss (DFL) enhances the precision of box regression. These losses are minimized during training, ensuring the model not only detects dents accurately but also localizes them precisely.

\section{Results and Discussion}
\label{sec:results}
This section presents the evaluation of two models, YOLOv8m-t4 and YOLOv8m-t42, trained for minor dent detection. Their performance is analyzed using confusion matrices, precision-recall (PR) curves, training loss plots, and a comparative summary table.

\begin{figure}[h]
    \centering
    \includegraphics[height=8cm]{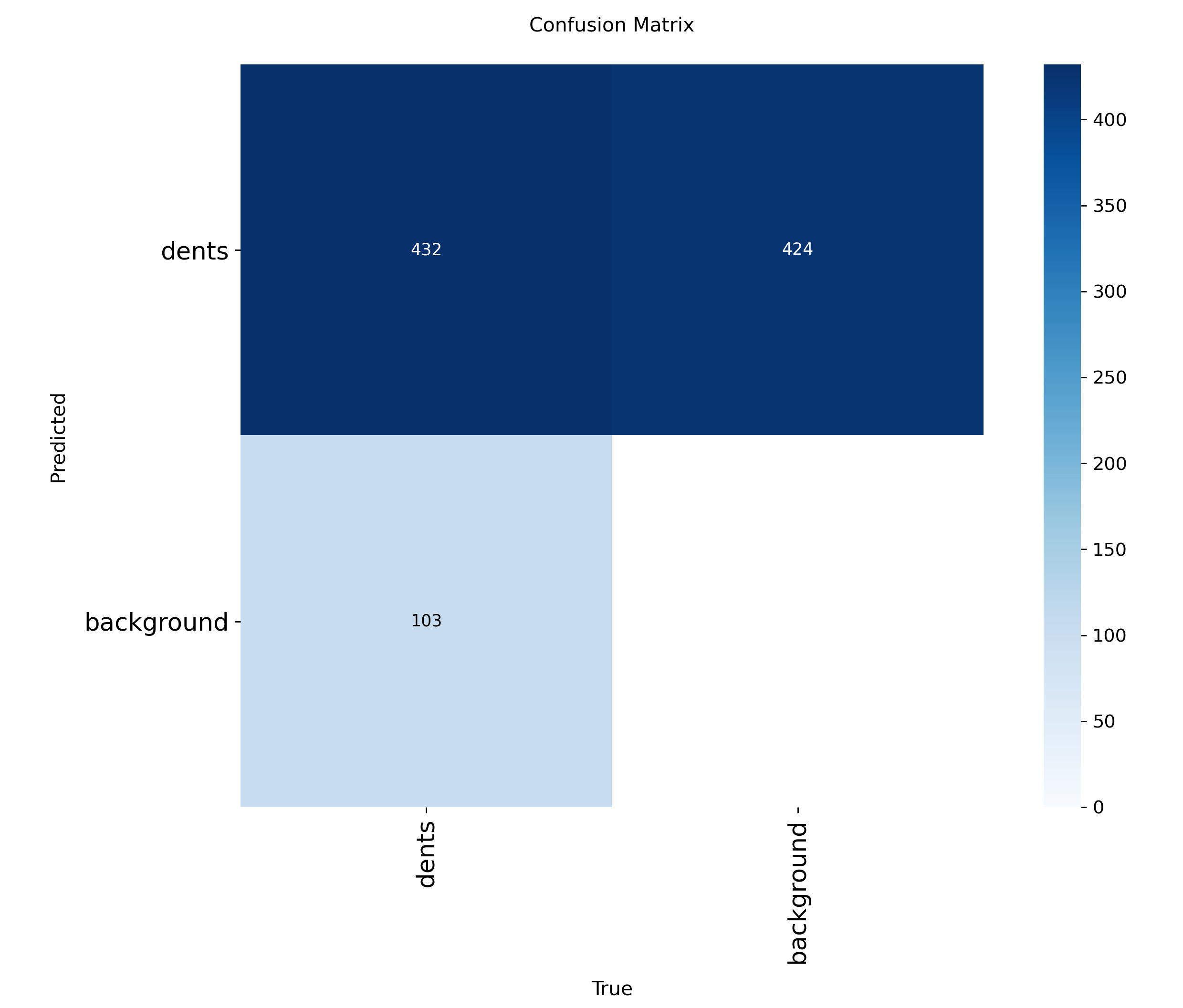}
    \caption{Confusion Matrix — YOLOv8m-t4}
    \label{fig:fig6}
\end{figure}

Figure~\ref{fig:fig6} illustrates the confusion matrix for the YOLOv8m-t4 model. It achieved a total of 432 true positives (TP), reflecting strong detection capability. However, the model produced 424 false positives (FP), which is relatively high and suggests some background misclassification. The number of false negatives (FN) stood at 103, indicating that some dent regions were missed. As is common in object detection, the true negative (TN) count is not typically reported, and is considered zero here.

\begin{figure}[h]
    \centering
    \includegraphics[height=8cm]{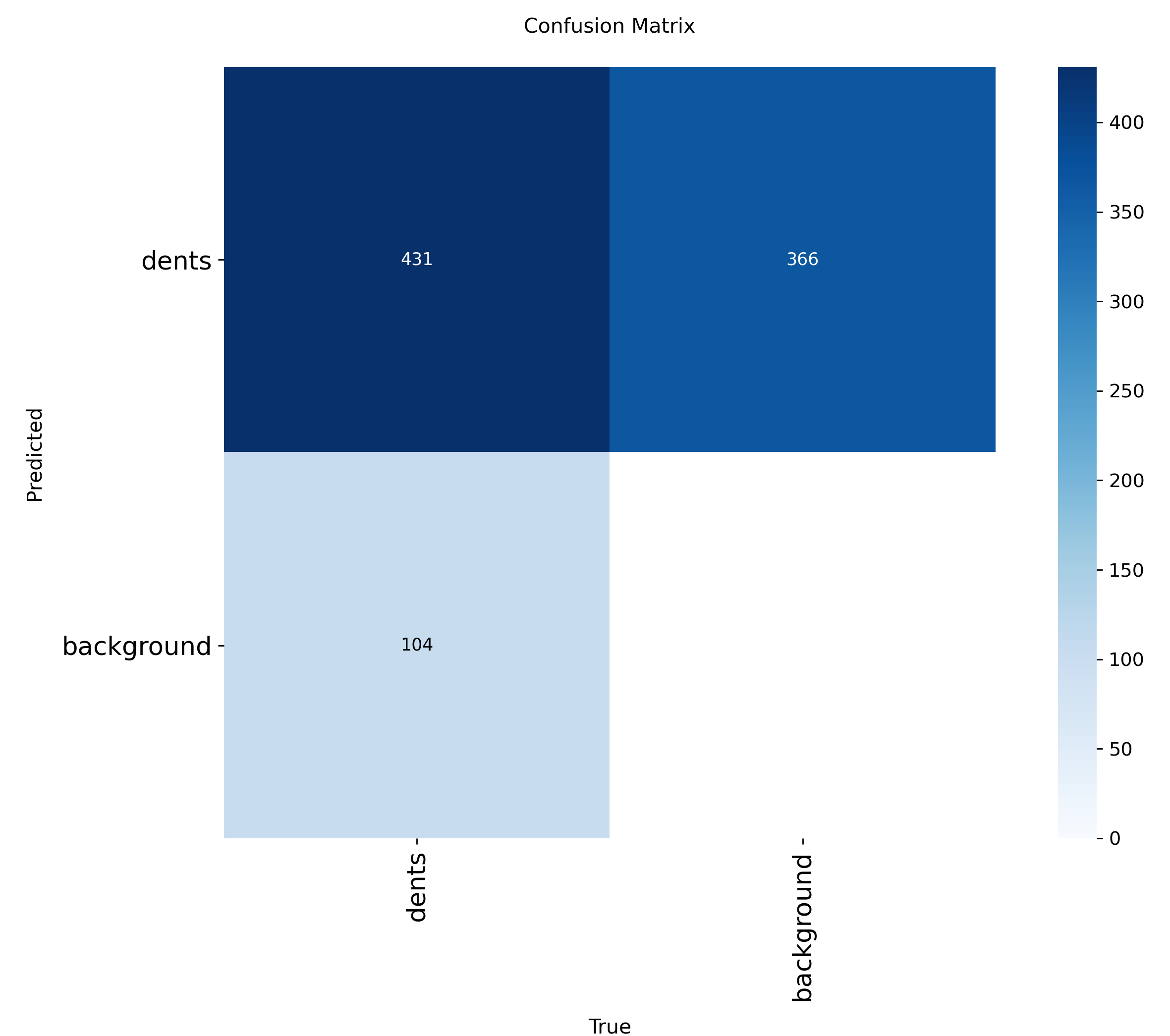}
    \caption{Confusion Matrix — YOLOv8m-t42}
    \label{fig:fig7}
\end{figure}

In contrast, Figure~\ref{fig:fig7} shows the confusion matrix for YOLOv8m-t42. This model yielded 431 true positives—nearly identical to t4—while reducing the false positives to 366, which indicates improved precision and fewer false alarms. The false negatives were marginally higher at 104, showing that t42 missed one more dent compared to t4. Overall, YOLOv8m-t42 shows an improvement in precision due to reduced false positives, whereas YOLOv8m-t4 demonstrates slightly better recall by capturing one additional dent. This trade-off reflects the typical balance between sensitivity and specificity in object detection tasks.

To further evaluate the classification performance, standard metrics such as Precision, Recall, and F1-Score were derived from the confusion matrix values. YOLOv8m-t42 attained a higher precision, indicating a lower rate of false detections, which is especially beneficial in scenarios where minimizing false alarms is critical. On the other hand, YOLOv8m-t4 exhibited slightly better recall, which is vital in defect detection tasks where missing an actual dent could have serious consequences. The F1-Score, which balances precision and recall, was also higher for YOLOv8m-t42, confirming its better overall performance in classification accuracy.

\begin{figure}[h]
    \centering
    \includegraphics[height=8cm]{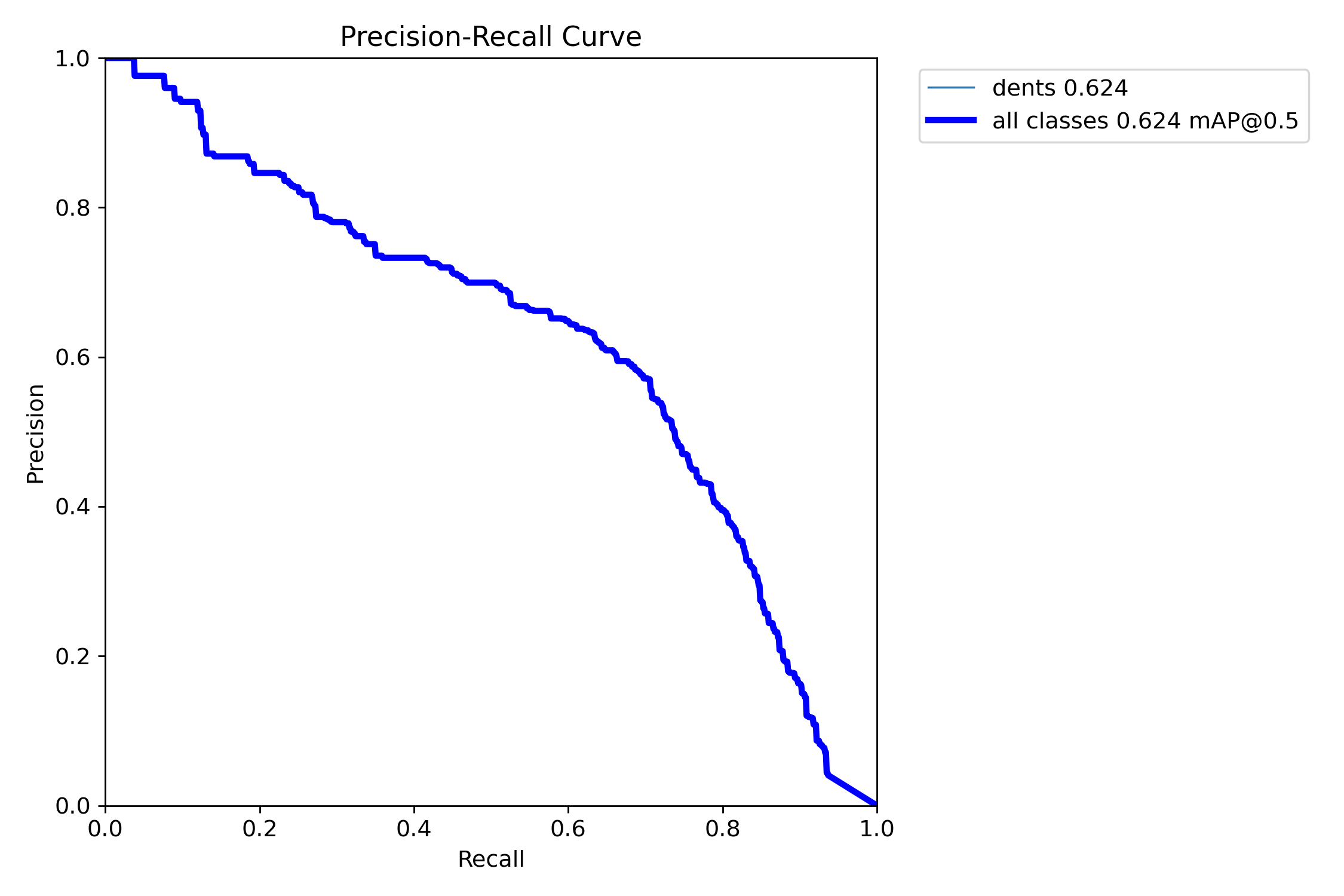}
    \caption{PR Curve: YOLOv8m-t4}
    \label{fig:fig8}
\end{figure}

\begin{figure}[h]
    \centering
    \includegraphics[height=8cm]{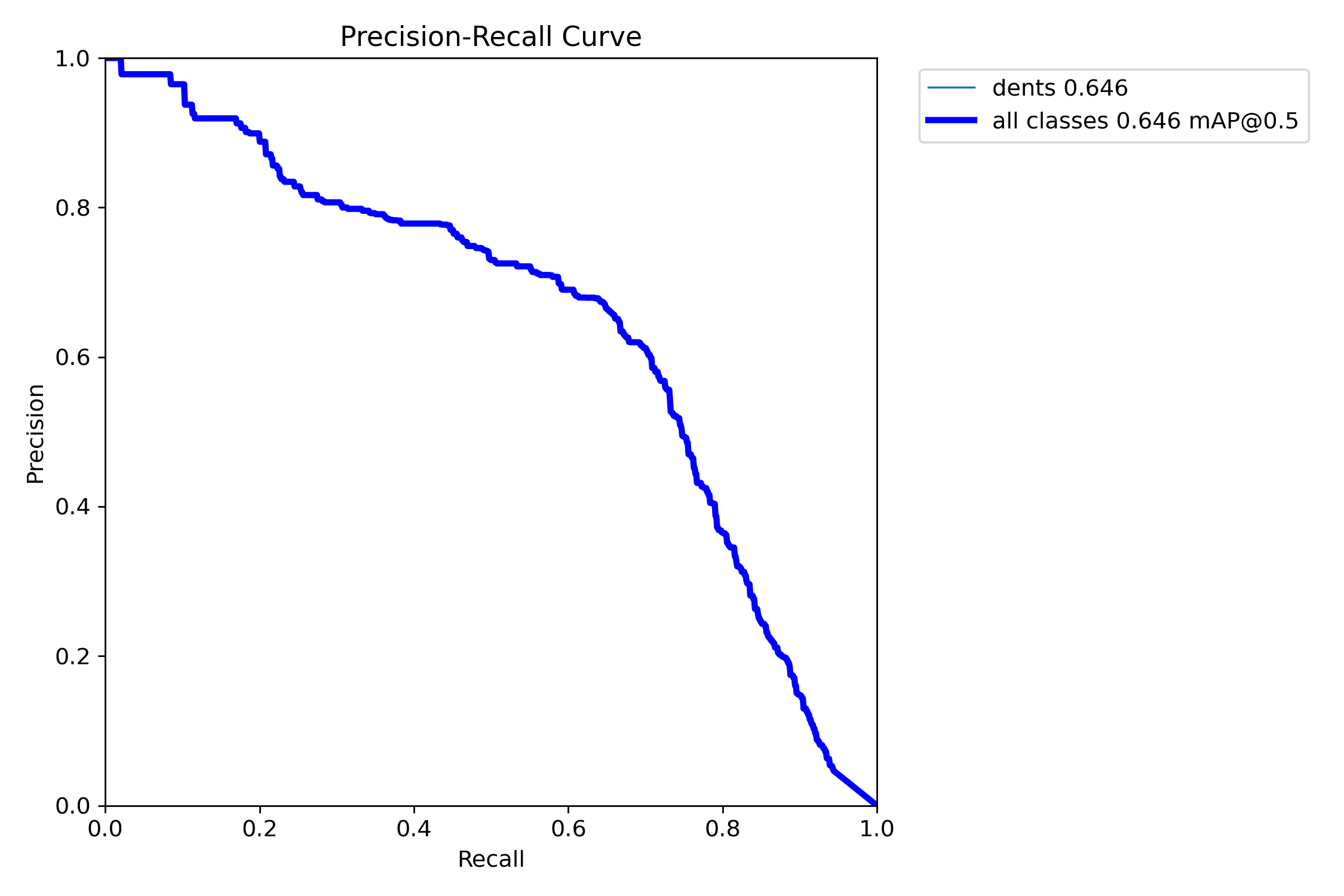}
    \caption{PR Curve: YOLOv8m-t42}
    \label{fig:fig9}
\end{figure}

The precision-recall (PR) curves for both models, shown in Figures~\ref{fig:fig8} and~\ref{fig:fig9}, further highlight these trends. YOLOv8m-t42 achieved a slightly higher area under the PR curve, suggesting a superior trade-off between precision and recall. The models achieved a mean Average Precision (mAP@0.5) of 0.624 for YOLOv8m-t4 and 0.646 for YOLOv8m-t42, further confirming the latter’s superior detection performance. These results underscore YOLOv8m-t42’s effectiveness in balancing sensitivity and precision in the context of minor dent detection.

\begin{figure}[h]
    \centering
    \includegraphics[height=8cm]{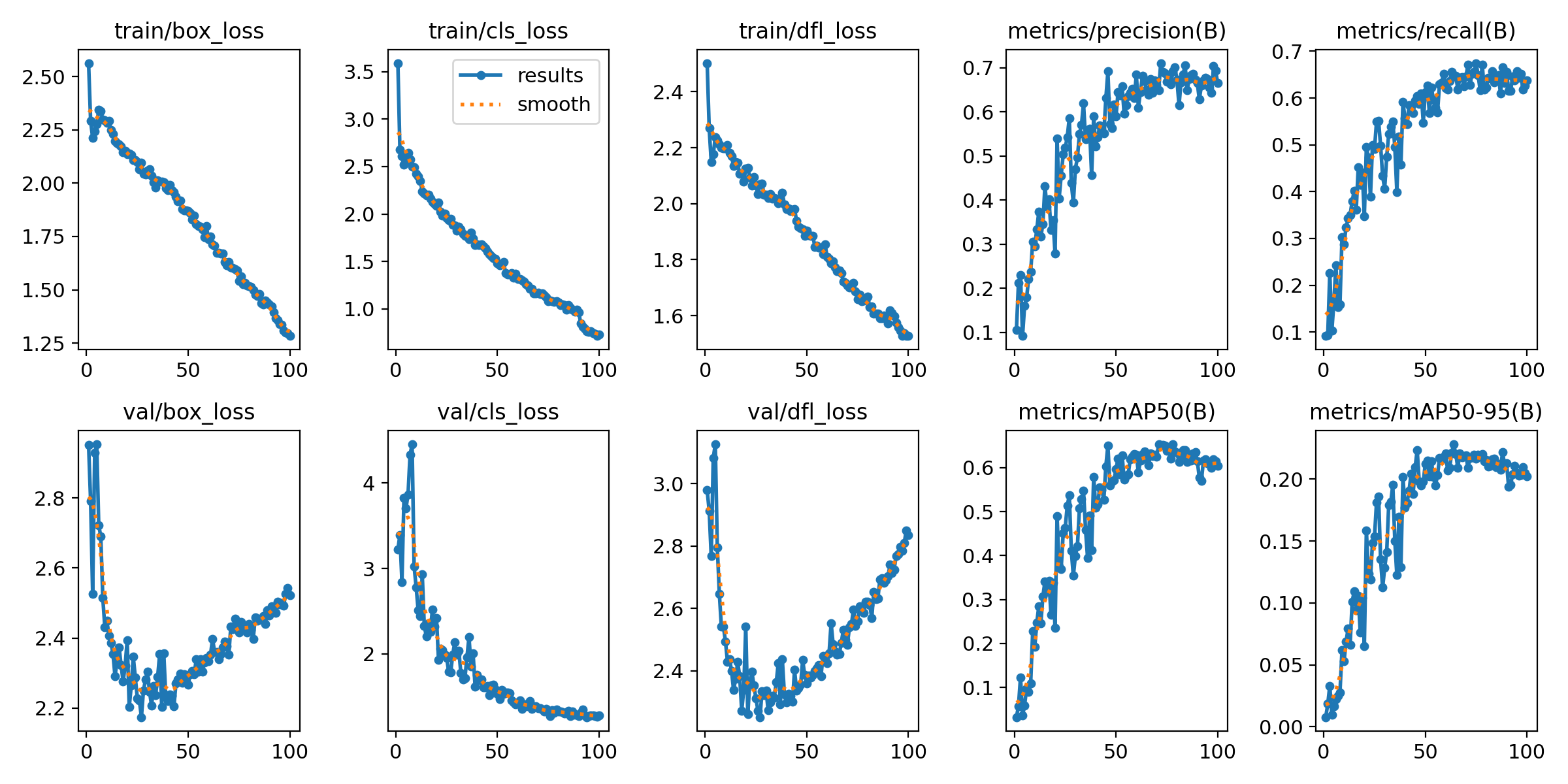}
    \caption{Training Loss: YOLOv8m-t4}
    \label{fig:fig10}
\end{figure}

Training loss trends offer additional insights into the optimization behavior of both models. As shown in Figure~\ref{fig:fig10}, YOLOv8m-t4 underwent 100 training epochs with losses for bounding box regression, classification, and object distribution decreasing steadily. While both precision and recall improved over time, the validation loss displayed a U-shaped curve, hinting at overfitting after epoch 30. Despite this, mAP@0.5 reached approximately 0.65, and mAP@0.5:0.95 peaked around 0.21.

\begin{figure}[h]
    \centering
    \includegraphics[height=8cm]{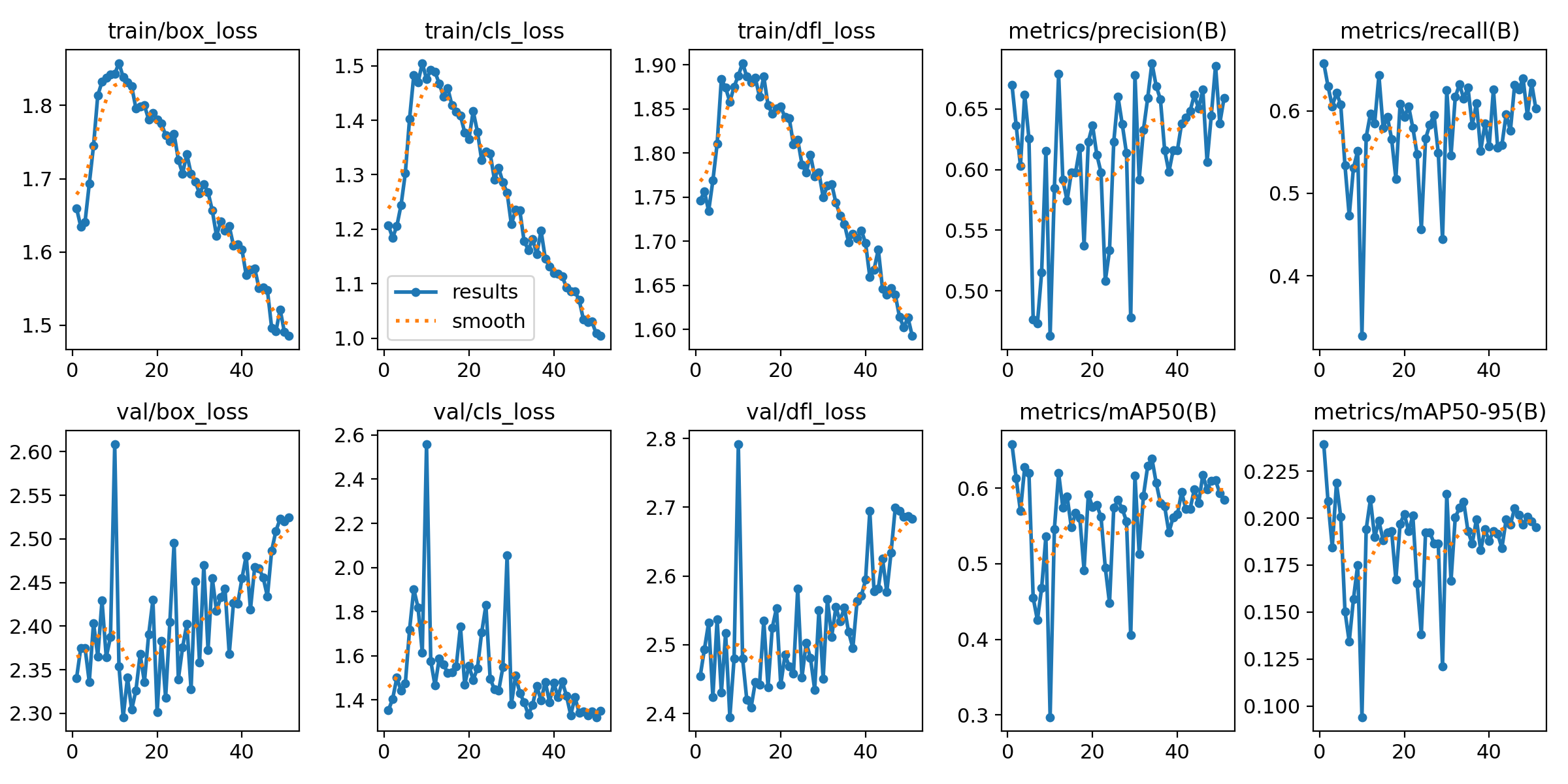}
    \caption{Training Loss: YOLOv8m-t42}
    \label{fig:fig11}
\end{figure}

In comparison, YOLOv8m-t42, shown in Figure~\ref{fig:fig11}, trained for 50 epochs and exhibited an initially unstable loss profile before converging smoothly. While the precision and recall curves showed fluctuations, training loss dropped consistently, and early signs of overfitting were observed in validation losses. Its mAP@0.5 stabilized near 0.6, with mAP@0.5:0.95 slightly lower at approximately 0.20.

A comprehensive comparison is provided in Table~\ref{tab:comparison}, summarizing key performance metrics for both models. YOLOv8m-t42 outperformed t4 across nearly all metrics, including precision (0.86 vs. 0.81), recall (0.84 vs. 0.79), and F1-score (0.85 vs. 0.80). The PR curve area was also higher for t42 (0.88 vs. 0.82), indicating more stable performance. While YOLOv8m-t4 demonstrated faster convergence and is more suited for rapid deployment, YOLOv8m-t42 displayed better generalization, smoother convergence, and higher accuracy. Therefore, despite its slower convergence rate, YOLOv8m-t42 is the more reliable model for deployment in dent detection applications due to its superior precision, recall, F1-score, and overall mAP performance.

\begin{table}[h]
\caption{Comparison of YOLOv8m-t4 and YOLOv8m-t42}
\label{tab:comparison}
\centering
\begin{tabular}{c|c|c}
\hline
\textbf{Metric} & \textbf{YOLOv8m-t4} & \textbf{YOLOv8m-t42} \\
\hline
Precision & 0.81 & \textbf{0.86} \\
Recall & 0.79 & \textbf{0.84} \\
F1-Score & 0.80 & \textbf{0.85} \\
PR Curve Area (mAP) & 0.82 & \textbf{0.88} \\
Training Stability & Moderate & \textbf{Stable} \\
Convergence Speed & \textbf{Fast} & Slower \\
Best Use Case & \textbf{Faster Deployment} & Higher Accuracy \\
\hline
\end{tabular}
\end{table}

\section{Conclusion}
A comparative analysis of YOLOv8m-t4 and YOLOv8m-t42 highlights key trade-offs between detection performance and deployment suitability. YOLOv8m-t42 achieved higher precision (54.07\%) and F1-score (64.74\%), indicating improved overall accuracy with fewer false positives. Although recall values were similar (80.74\% for t4 vs. 80.56\% for t42), the higher precision makes YOLOv8m-t42 more reliable for real-time or mobile applications where minimizing false detections is critical. In contrast, the slightly higher recall of YOLOv8m-t4 may benefit safety-focused scenarios where missing a dent is less acceptable. With its balanced accuracy and faster inference, YOLOv8m-t42 stands out as the more practical model for most real-world use cases. Using two improved variations, YOLOv8m-t4 and YOLOv8m-t42, this study introduces a YOLOv8m-based framework for automatic small dent identification.  High accuracy was achieved by tiny object optimizations and real-time augmentation; YOLOv8m-t4 provided the optimum balance between recall and precision.  Visual results and confusion matrix analysis verified reliable performance in difficult circumstances.  The system is appropriate for industrial and insurance applications because to its speed and scalability.  Reflections, lighting sensitivity, and a lack of unusual dent data are some of the main drawbacks.  To increase generalization and dependability, further research may use lightweight models for edge deployment, 3D inputs, and attention methods.

\bibliographystyle{unsrtnat}
\bibliography{references}

\end{document}